\pdfoutput=1

\documentclass[11pt]{article}

\usepackage{emnlp2021}
\usepackage{times}
\usepackage{latexsym}
\usepackage{graphicx}
\usepackage{bbding}
\usepackage{epsfig}
\usepackage{graphicx}
\usepackage{amsmath}
\usepackage{amssymb}
\usepackage{threeparttable}
\usepackage[T1]{fontenc}

\usepackage[utf8]{inputenc}

\usepackage{microtype}

\title{R$^3$Net:Relation-embedded Representation Reconstruction Network \\ for Change Captioning }

\author{Yunbin Tu\textsuperscript{1}*, Liang Li\textsuperscript{2}$^{\dagger}$, \textbf{Chenggang Yan\textsuperscript{3}}, \textbf{Shengxiang Gao\textsuperscript{1},} \textbf{Zhengtao Yu\textsuperscript{1}$^{\dagger}$}  \\ \textsuperscript{1}Yunnan Key Laboratory of Artificial Intelligence, \\
 Kunming University of Science and Technology \\ \textsuperscript{2} Key Lab of Intell. Info. Process., Inst. of Comput. Tech., Chinese Academy of Sciences \\
 \textsuperscript{3} Intelligent Information Processing Laboratory, Hangzhou Dianzi University \\ 
 \texttt{tuyunbin1995@foxmail.com,} 
 \texttt{liang.li@ict.ac.cn}}
 \renewcommand{\thefootnote}{}
\begin{document}
\maketitle
\footnote{* This work was done at VIPL research group, CAS.}

\footnote{$^{\dagger}$ Corresponding author}
\renewcommand{\thefootnote}{1}
\begin{abstract}
Change captioning is to use a natural language sentence to describe the fine-grained disagreement between two similar images. Viewpoint change is the most typical distractor in this task, because it changes the scale and location of the objects and overwhelms the representation of real change. In this paper, we propose a Relation-embedded Representation Reconstruction Network (R$^3$Net) to explicitly distinguish the real change from the large amount of clutter and irrelevant changes. Specifically, a relation-embedded module is first devised to explore potential changed objects in the large amount of clutter. Then, based on the semantic similarities of corresponding locations in the two images, a representation reconstruction module (RRM) is designed to learn the  reconstruction representation and further model the difference representation. Besides, we introduce a syntactic skeleton predictor (SSP) to enhance the semantic interaction between change localization and caption generation. Extensive experiments show that the proposed method achieves the state-of-the-art results on two public datasets \footnote{The code of this paper has been made publicly available at \url{https://github.com/tuyunbin/R3Net}.}.

\end{abstract}

\section{Introduction}

Change captioning aims to generate a natural language sentence to detail what has changed in a pair of similar images.
It has many practical applications, such as assisted surveillance, medical imaging, and computer assisted tracking of changes in media assets \cite{DBLP:conf/emnlp/JhamtaniB18,DBLP:conf/acl/TuYLLGYY21}.

Different from single-image captioning \cite{kim2019image,jiang2019reo,fisch2020capwap}, change captioning addresses two-image captioning, which requires not only to understand both image content, but also to describe their disagreement. As the pioneer work, Jhamtani \emph{et al.} \cite{DBLP:conf/emnlp/JhamtaniB18} described semantic changes between mostly well-aligned image pairs with underlying illumination changes from surveillance cameras. However, they did not consider viewpoint changes that
often happen in a dynamic world, and image pairs cannot be mostly well aligned in this case. Hence, feature shift between two unaligned images will adversely affect the learning of difference representation. To make this task more practical, recent works \cite{Park2019RobustCC,DBLP:conf/eccv/ShiYGJ020} proposed to address change captioning in the presence of viewpoint changes. 

\begin{figure}[t]
\centering
\includegraphics[width=0.45\textwidth]{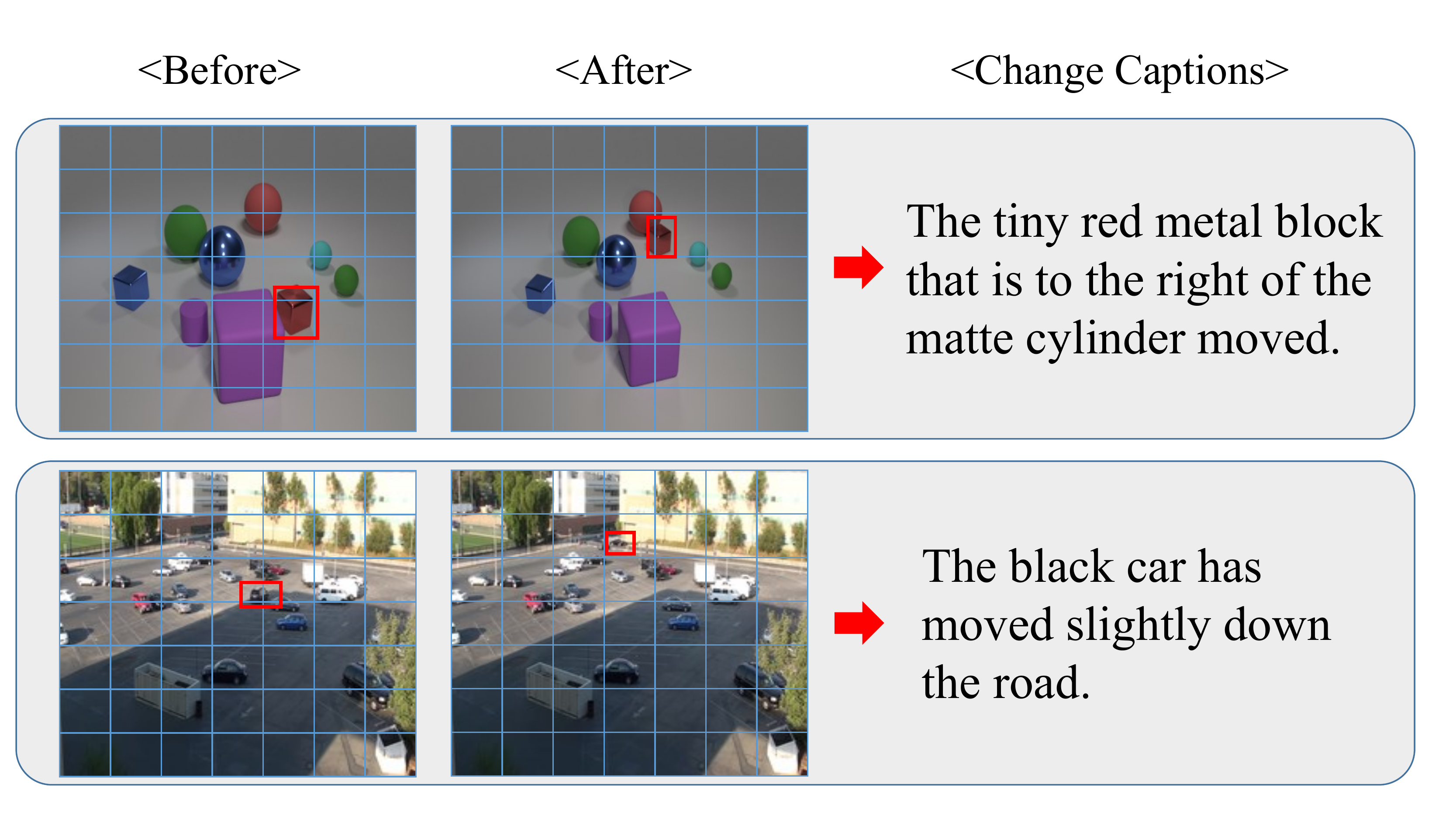} 
\caption{Two examples of change captioning about an object move. The first example shows that the viewpoint changes the scale and location of the objects in the ``after'' image; the second example shows mostly well-aligned a pair of images with underlying illumination changes from surveillance cameras. }
\label{fig1}
\end{figure}

Despite the progress, there are some limitations for the above state-of-the-art methods when modeling the difference representation. First, the object information of each image is only learned at feature-level, and this is difficult to discriminate fine-grained difference when changed object is too tiny and surrounded by the large amount of clutter, as shown in Figure \ref{fig1}.
Actually, when an object moved, its semantic relations with surrounding objects would change as well, and this can help explore the fine-grained change. Thus, it is important to model the difference representation at both feature and relation levels. 
Second, directly applying subtraction between a pair of unaligned images \cite{Park2019RobustCC} may learn the difference representation with much noise, because viewpoint changes the scale and location of the objects. However, we can observe that those unchanged objects are still in the approximate locations. Hence, it is beneficial to reveal the unchanged representation and further model the difference representation based on the semantic similarities in the corresponding locations of two images.

In this paper, we propose a Relation-embedded Representation Reconstruction Network (R$^3$Net) to handle viewpoint changes and model the fine-grained difference representation between two images in the process of representation reconstruction. Concretely, for ``before'' and ``after'' images, the relation-embedded module respectively performs semantic relation reasoning among object features via the self-attention mechanism. This can enhance the fine-grained representation ability of original object features. To model the difference representation, a representation reconstruction module (RRM) is designed, where a ``shadow'' representation (``after'' or ``before'') is used to reconstruct a ``source'' representation (``before'' or ``after''). The RRM first leverages every location in the ``source'' to stimulate the corresponding locations in the ``shadow'' to judge their semantic similarities, \emph{i.e.},  ``response signals''.
Further, under the guidance of the signals, the RRM picks out the unchanged features from the ``shadow'' as the ``reconstruction'' representation. The ``difference'' representation is computed with the changed features between the ``source'' and ``reconstruction''. 
Next, a dual change localizer is devised to use the representation of difference  as the query to localize the changed object feature on the ``before'' and ``after'', respectively. Finally, the localized features are fed into an attention-based caption decoder for caption generation.

Besides, we introduce a Syntactic Skeleton Predictor (SSP) to enhance the semantic interaction between change localization and caption generation. As observed in Figure \ref{fig1}, a caption mainly consists of a set of nouns, adjectives, and verbs. These words can convey main information of the changed object and its surrounding references, called \emph{syntactic skeletons}. The skeletons, which are predicted based on a global semantic representation derived from the R$^3$Net, can supervise the modeling of difference representation and provide the decoder with high-level semantic cues about change type. This makes
the learned difference representation more relevant to the target words and enhances the quality of generated sentences.

The main contributions of this paper are as follows: (1) We propose the R$^3$Net to learn the fine-grained change from the large amount of clutter and overcome viewpoint changes by embedding semantic relations into object features and performing representation reconstruction with respect to the two images. (2) The SSP is introduced to enhance the semantic interaction between change localization and caption generation via predicting a set of syntactic skeletons based on a global semantic representation derived from the R$^3$Net.  (3) Extensive experiments show that the proposed method outperforms the state-of-the-art approaches by a large margin on two public datasets.

\begin{figure*}[h]
\centering
\includegraphics[width=1.0\textwidth]{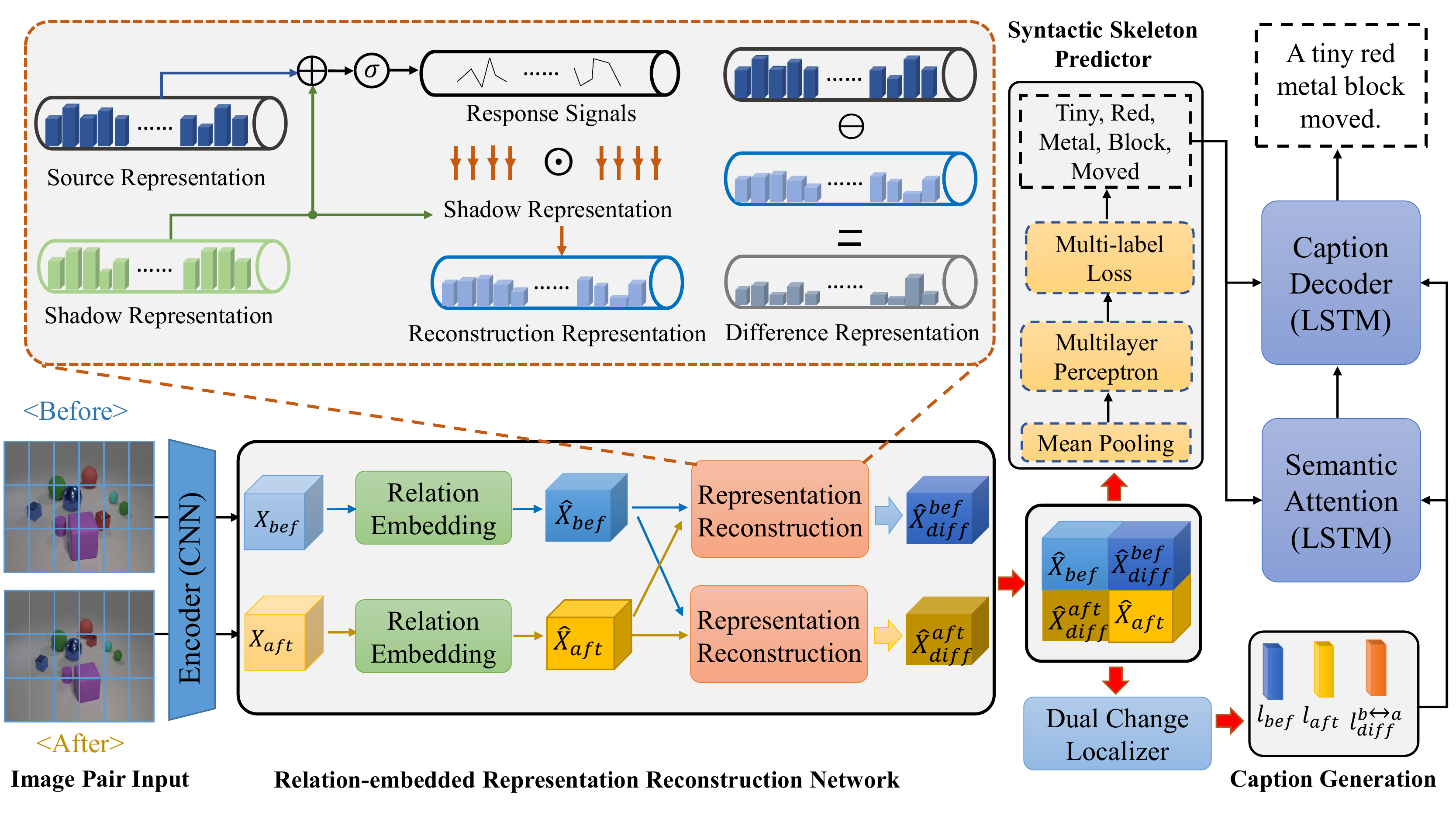} 
  \caption{The architecture of the proposed method, consisting of a relation-embedded representation reconstruction network, a syntactic skeleton predictor, a dual change localizer and an attention-based caption decoder.}
\label{fig2}
\end{figure*}

\section{Related Work}
\textbf{Change Captioning.} Captioning the change in the existence of viewpoint changes is a novel task in the vision-language community \cite{zhang2017task,tu2017video,deng2021syntax,li2020diverter,liu2020ir}. As the first work, DUDA \cite{Park2019RobustCC} directly applied subtraction between two images to capture their semantic difference. However, due to viewpoint changes, direct subtraction between an unaligned image pair cannot reliably model the correct change \cite{DBLP:conf/eccv/ShiYGJ020}. Later, M-VAM \cite{DBLP:conf/eccv/ShiYGJ020} proposed to measure the feature similarity across different regions in an image pair and find the most matched regions as unchanged parts. However, since there are a lot of similar objects, cross-region searching will face the risk of matching the query region with a similar but incorrect region, impacting subsequent change localization. In contrast, in our representation reconstruction network, the prediction of unchanged and changed features are based on the semantic similarities of the corresponding locations in two images. This can avoid the risk of reconstructing ``source'' with incorrect parts from ``shadow''.

\textbf{Skeleton Prediction in Captioning.} Syntactic skeletons can provide the high-level semantic cues (e.g., attribute, class) about objects, so they are widely used in image/video captioning works.
These methods either used skeletons as main information to generate captions \cite{fang2015captions,gan2017semantic,dai2018neural} or leveraged them to bridge the semantic gap between vision and language \cite{gao2020fused,tu2020enhancing}. Although the skeletons played different roles in the above methods, the common point was that they only represent basic information of objects in images or videos. Different from them, besides basic information, we try to use skeletons to capture the changed information among objects.



\section{Methodology}
As shown in Figure \ref{fig2}, the architecture of our method consists of four main parts: (1) a relation-embedded representation reconstruction network (R$^3$Net) to learn the fine-grained change in the presence of viewpoint changes; (2) a dual change localizer to focus on the specific change in a pair of images; (3) a syntactic skeleton predictor (SSP) to learn syntactic skeletons based on a global semantic representation derived from  the R$^3$Net; (4) an attention-based caption decoder to describe the change under the guidance of the learned skeletons.

\subsection{Relation-embedded Representation Reconstruction Network}

\subsubsection{Relation-embedded Module}
We first exploit a pre-trained CNN model to extract the object-level features $X_{bef}$ and $X_{aft}$ for a pair of ``before'' and ``after'' images, where $X_{i}\in \mathbb{R}^{C \times H \times W}$ and C, H, W indicate the number of channels, height, and width. However, only utilizing these independent features is difficult to distinguish fine-grained change from the large amount of clutter (similar objects). And related works \cite{wu2019connective,huang2020aligned,yin2020novel} have shown that capturing semantic relations among objects is useful for a thorough understanding of an image.

Motivated by this, we devise a relation-embedded (R$_{emb}$) module based on the self-attention mechanism \cite{vaswani2017attention} to implicitly learn semantic relations among objects in each image. Specifically, we first reshape $X_{i}\in \mathbb{R}^{C \times H \times W}$ to $X_{i}\in \mathbb{R}^{N \times C}$ ($N=HW$), where $i\in (bef, aft)$. Then, the semantic relations are embedded into independent object features of each image based on the scaled dot-product attention:
\begin{equation}
\mathrm{R}_{emb}(Q, K, V)=\operatorname{softmax}\left(\frac{Q K^{T}}{\sqrt{d_{k}}}\right) V,
\end{equation}
where the quires, keys and values are the projections of the object features in $X_{i}$ and $i\in (bef, aft)$:
\begin{equation}
(Q,K,V) = \left(X_{i} W_{i}^{Q}, X_{i} W_{i}^{K}, X_{i} W_{i}^{V}\right).
\end{equation}
Thus, $X_{bef}$ and $X_{aft}$ are updated to $\hat X_{bef}$ and $\hat X_{aft}$, respectively:
\begin{equation}
\begin{array}{l}
\hat{X}_{b e f}=\mathrm{R}_{e m b}\left(X_{b e f}, X_{b e f}, X_{b e f}\right), \\
\hat{X}_{a f t}=\mathrm{R}_{e m b}\left(X_{a f t}, X_{a f t}, X_{a f t}\right).
\end{array}
\end{equation}
When the model fully understands each image content, it can
better capture the fine-grained difference between the image pair in the subsequent representation reconstruction.


\subsubsection{Representation Reconstruction Module}
The state-of-the-art method \cite{Park2019RobustCC} applied direct subtraction between a pair of unaligned images, which is prone to capture the difference with noise in the presence of viewpoint changes. 

To distinguish semantic change from viewpoint changes, a representation reconstruction module (RRM) is proposed, where the inputs are a ``source'' representation $\hat X_{p}\in \mathbb{R}^{N \times C}$ and a ``shadow'' representation $\hat X_{s}\in \mathbb{R}^{N \times C}$. Concretely, first, we exploit each location of  $\hat X_p$ to stimulate the corresponding location of $\hat X_s$. The response degrees of all locations in $\hat X_s$ are regarded as the response signals $\alpha$ that measure the semantic similarities between corresponding locations in two images :
\begin{equation}
\alpha=\operatorname{Sigmoid}\left(\hat X_{p} W_{p}+\hat X_{s} W_{s} + b_s\right),
\end{equation}
where $W_p$, $W_{s}\in \mathbb{R}^{C \times C}$ and $b_{s}\in \mathbb{R}^{C}$. 
Second, we use $\hat X_s$ to reconstruct $\hat X_p$ under the guidance of the response signals $\alpha$:
\begin{equation}
\Tilde X_{p}=\alpha \odot \hat X_{s},
\end{equation}
where $\Tilde X_{p}\in \mathbb{R}^{N \times C}$ is the ``reconstruction'' representation, which represent unchanged features with respect to ``source''.
Finally, the ``difference'' representation is captured by subtracting ``reconstruction'' $\Tilde X_{p}$ from ``source'' $\hat X_p$:
\begin{equation}
\hat X_{d i f f}=\hat X_{p}-\Tilde X_{p}.
\end{equation}

Since the predicted unchanged and changed features in uni-directional reconstruction are only with respect to one kind of ``source'' representation (e.g., ``before''), the model cannot predict the changed feature when it is not in the ``source''. 
For an efficient model, it should capture all underlying changes with respect to both images. To this end, we extend the RRM from uni-direction to bi-direction. Specifically, we first use the ``before'' as ``source'' to predict unchanged and changed features, and then use the ``after'' as ``source'' to do so. Thus, the ``reconstruction'' and ``difference'' w.r.t. the ``before'' and ``after'' are formulated as:
\begin{equation}
\begin{array}{l}
\Tilde X_{p}^{bef}, \hat X_{diff}^{bef}=\operatorname{RRM}\left(\hat X_{p}^{bef} , \hat X_{s}^{aft}\right), \\
\Tilde X_{p}^{aft}, \hat X_{diff}^{aft}=\operatorname{RRM}\left(\hat X_{p}^{aft} , \hat X_{s}^{bef}\right).
\end{array}
\end{equation}
Finally, we obtain a bi-directional difference representation by a fully-connected layer:
\begin{equation}
\hat X_{d i f f}=\operatorname{ReLU}\left(\left[\hat X_{diff}^{bef} ; \hat X_{diff}^{aft}\right] W_{f} + b_{f}\right).
\end{equation}

\subsection{Dual Change localizer}
When the bi-directional difference representation $\hat X_{diff}$ is computed, we exploit it as the query to localize the changed feature in $\hat X_{bef}$ and $\hat X_{aft}$, respectively. Specifically, the dual change localizer first predicts two separate attention maps $a_{bef}$ and $a_{aft}$:
\begin{equation}
\begin{array}{c}
X_{\text {bef }}^{\prime}=\left[\hat X_{\text {bef }} ; \hat X_{\text {diff }}\right], 
X_{\text {aft }}^{\prime}=\left[\hat X_{\text {aft }} ; \hat X_{\text {diff }}\right], \\
a_{\text {bef }}=\sigma\left(\operatorname{conv}_{2}\left(\operatorname{ReLU}\left(\operatorname{conv}_{1}\left(X_{\text {bef }}^{\prime}\right)\right)\right)\right), \\
a_{\text {aft }}=\sigma\left(\operatorname{conv}_{2}\left(\operatorname{ReLU}\left(\operatorname{conv}_{1}\left(X_{\text {aft }}^{\prime}\right)\right)\right)\right),
\end{array}
\end{equation}
where [;], conv, and $\sigma$ denote concatenation, convolutional layer, and sigmoid activation function, respectively. Then, the changed features $l_{bef}$ and $l_{aft}$ are localized via applying $a_{bef}$ and $a_{aft}$ to $\hat X_{bef}$ and $\hat X_{aft}$:
\begin{equation}
\begin{aligned}
l_{\text {bef }} &=\sum_{H, W} a_{\text {bef }} \odot \hat X_{\text {bef }}, l_{\text {bef }} \in \mathbb{R}^{C}, \\
l_{\text {aft }} &=\sum_{H, W} a_{\text {aft }} \odot \hat X_{\text {aft }}, l_{\text {aft }} \in \mathbb{R}^{C}.
\end{aligned}
\end{equation}
Finally, we compute the local difference feature w.r.t. both $l_{bef}$ and $l_{aft}$ from two directions:
\begin{equation}
\begin{array}{c}
l_{d i f f}^{b \rightarrow a}=l_{b e f}-l_{a f t}, \quad
l_{d i f f}^{a \rightarrow b}=l_{a f t}-l_{b e f}, \\
l_{d i f f}^{b \leftrightarrow a}=\operatorname{ReLU} \left(W_{d}\left[l_{d i f f}^{b \rightarrow a} ; l_{d i f f}^{a \rightarrow b}\right] + b_d\right).
\end{array}
\end{equation}

\subsection{Syntactic Skeleton Predictor}
A syntactic skeleton predictor (SSP) is introduced to learn a set of syntactic skeletons based on the outputs derived from the R$^3$Net. The predicted skeletons can provide the caption decoder with high-level semantic cues about changed objects and supervise the modeling of difference representation. This aims to enhance the semantic interaction between change localization and caption generation. Inspired by \cite{gao2020fused,gan2017semantic}, we treat this problem as a multi-label classification task. Suppose there are $N$ training image pairs, and ${y}_{j}=\left[y_{j 1}, \ldots, y_{j K}\right] \in\{0,1\}^{K}$ is the label vector of the $j$-th image pair, where $y_{jk}=1$ if the image pair is annotated with the skeleton $k$, and $y_{jk}=0$ otherwise. 

Specifically, first,  we apply a mean-pooling layer over the concatenated semantic representations of $\hat X_{bef}$, $\hat X_{aft}$, and $ \hat X_{diff}$ to obtain a global semantic representation ${S}_{j}$:
\begin{equation}
S_{j}=\frac{1}{H, W} \sum_{H, W}\left[\hat X_{\text {bef }} ; \hat X_{\text {aft }} ; \hat X_{\text {diff }}\right].
\end{equation}
Then, the probability scores $p_{j}$ of all syntactic skeletons for $j$-th image pair is computed by:
\begin{equation}
\left.p_{j}=\operatorname{sigmoid}\left(U_{\mathrm{g}} \operatorname{ReLU} (W_{g} S_{j} \right)+b_g\right),
\end{equation}
where ${p}_{j}=\left[p_{j 1}, \ldots, p_{j K}\right]$ denotes the probability scores of $K$ skeletons in $j$-th image pair.
To maximize the probability scores of syntactic skeletons, we use the multi-label loss to optimize the SSP. It can be formulated as:
\begin{equation}
\begin{array}{r}
L_{s}=-\frac{1}{N} \sum_{j=1}^{N} \sum_{k=1}^{K}\left(y_{j k} \log p_{j k}+\right. \\
\left(1-y_{j k}\right) \log \left(1-p_{j k})\right),
\end{array}
\label{multi}
\end{equation}
where $N$ and $K$ indicate the number of all training samples and annotated skeletons of an image pair. The loss can be considered as the supervision signal to regularize the learning of difference representation in the R$^3$Net.


\subsection{Skeleton-guided Caption generation}
Since the predicted skeletons are the explicit semantic concepts of the changed object and its surrounding references, the captions are generated under the guidance of them. Specifically, first, the predicted probability scores $p_{j}$ are embedded as a skeleton feature $E[p_j]$:
\begin{equation}
E\left[p_{j}\right]=\operatorname{ReLU}\left(W_{q}\left(E_{q} p_{j} \right) + b_q \right),
\end{equation}
where $E_q \in \mathbb{R}^{k \times M}$ is a skeleton embedding matrix and $M$ is the dimension of the skeleton feature. $W_{q}\in \mathbb{R}^{M \times M}$ and $b_{q}\in \mathbb{R}^{M}$ are the parameters to be learned. 
Then, we exploit a semantic attention module to focus on the key semantic feature from $l_{bef}$, $l_{aft}$, and $l_{d i f f}^{b \leftrightarrow a}$, which is relevant to the target word:
\begin{equation}
l_{\mathrm{dyn}}^{(t)}=\sum_{i} \beta_{i}^{(t)} l_{i},
\end{equation}
where $i\in (bef, aft, diff)$. $\beta_{i}^{(t)}$ is
computed by an attention LSTM$_a$ under the guidance of the predicted skeleton feature $E[p_j]$:
\begin{equation}
\begin{array}{c}
v=\operatorname{ReLU}\left(W_{a_{1}}\left[l_{\text {bef }} ; l_{d i f f}^{b \leftrightarrow a} ; l_{\text {aft }}\right]+b_{a_{1}}\right), \\
\qquad{u^{(t)}=\left[v; E\left[p_j\right]; h_{c}^{(t-1)}\right]}, \\
h_{a}^{(t)}=\operatorname{LSTM}_{a}\left(h_{a}^{(t)} \mid u^{(t)}, h_{a}^{(0: t-1)}\right), \\
\beta^{(t)} \sim \operatorname{Softmax}\left(W_{a_{2}} h_{a}^{(t)}+b_{a_{2}}\right).
\end{array}
\end{equation}
where $W_{a_{1}}, b_{a_{1}}, W_{a_{2}},$ and $b_{a_{2}}$ are learnable parameters. $h_{a}^{(*)}$ and $h_{c}^{(*)}$ are hidden states of the attention module LSTM$_a$ and the caption decoder LSTM$_c$, respectively.

Finally, the caption generation process is also guided by the predicted skeleton feature. We feed it, attended visual feature, and the previous word embedding to the caption decoder LSTM$_c$ to predict a series of distributions over the next word:
\begin{equation}
\begin{array}{c}
c^{(t)}=\left[E\left[w_{t-1}\right]; E\left[p_j\right]; l_{\mathrm{dyn}}^{(t)}\right],\\
h_{c}^{(t)}=\operatorname{LSTM}_{c}\left(h_{c}^{(t)} \mid c^{(t)}, h_{c}^{(0: t-1)}\right), \\
w_{t} \sim \operatorname{Softmax}\left(W_{c} h_{c}^{(t)}+b_{c}\right),
\end{array}
\end{equation}
where $E$ is a word embedding matrix; $W_{c}$ and $b_{c}$ are learnable parameters.

\subsection{Joint Training}
We jointly train the caption decoder and SSP in an end-to-end
manner. 
For the SSP, the multi-label loss is minimized by the Eq. (\ref{multi}). 
For the decoder, given the target ground-truth words $\left(w_{1}, \ldots, w_{m}\right)$, we minimize its negative log-likelihood loss:
\begin{equation}
L_{cap}(\theta_c)=-\sum_{t=1}^{m} \log p\left(w_{t} \mid w_{<t}; \theta_c\right),
\end{equation}
where $\theta_c$ are the parameters of the decoder and $m$ is the length of the caption. The final loss function is optimized as follows:
\begin{equation}
L(\theta)=L_{cap}+\lambda*L_{s},
\end{equation}
where the hyper-parameter $\lambda$ is to seek a trade-off between the decoder and SSP.

\begin{table*}[h]
\centering
\begin{threeparttable}
  \caption{Ablation studies on CLEVR-Change in terms of total performance.}
    \label{table1}
\begin{tabular}{c|c|c|c|c|c}
\hline
            & \multicolumn{5}{c}{Total}                                                     \\ \hline
Method     & BLEU-4           & METEOR           & ROUGE-L             & CIDEr             & SPICE             \\ \hline
Baseline        & 53.1          & 37.6          & 70.8          & 115.6          & 31.3          \\ \hline
RRM         & 53.5         & 39.2        & 72.3            & 119.2          & 32.5          \\ \hline
R$^3$Net         & 54.2           & 39.4          & 72.7            & 122.3          & \textbf{32.6}          \\ \hline
R$^3$Net+SSP & \textbf{54.7}  & \textbf{39.8} & \textbf{73.1}  & \textbf{123.0} & \textbf{32.6}   \\ \hline   
\end{tabular}
\end{threeparttable}
\end{table*}

\begin{table*}[h]
\centering
\begin{threeparttable}
  \caption{Ablation studies on CLEVR-Change in terms of two settings, where B-4, M, R, C, and S are short for BLEU-4, METEOR, ROUGE-L, CIDEr, and SPICE, respectively.}
    \label{table2}
\begin{tabular}{c|c|c|c|c|c|c|c|c|c|c}
\hline
            & \multicolumn{5}{c|}{Scene Change}                                              & \multicolumn{5}{c}{None-scene Change}                                                \\ \hline
Method      & B-4           & M             & R             & C              & S             & B-4           & M             & R             & C              & S             \\ \hline
Baseline    & 51.0          & 33.3          & 65.7          & 102.4          & 28.0          & 61.0          & 49.9          & 75.8         & 114.3          & 34.5          \\ \hline
RRM         & 51.8          & 35.7          & 69.0          & 110.1          & 30.4          & 60.0          & 49.6          & 75.6          & 115.0          & 34.5          \\ \hline
R$^3$Net         & 52.5          & 36.0          & 69.5          & 114.8          & \textbf{30.5}         & \textbf{62.0}          & 50.0          & 75.9          & 116.3         &  \textbf{34.8}          \\ \hline

R$^3$Net+SSP & \textbf{52.7} & \textbf{36.2} & \textbf{69.8} & \textbf{116.6} & 30.3 & 61.9 & \textbf{50.5}          & \textbf{76.4}          & \textbf{116.4} & \textbf{34.8} \\ \hline
\end{tabular}
\end{threeparttable}
\end{table*}

\section{Experiments}
\subsection{Datasets and Evaluation Metrics}
\textbf{CLEVR-Change} dataset \cite{Park2019RobustCC} is a large-scale dataset with a set of basic geometry objects, which consists of 79,606 image pairs and 493,735 captions. The change types can be categorized into six cases, \emph{i.e.}, ``Color'', ``Texture'', ``Add'', ``Drop'', `'Move'' and ``Distractors (e.g., viewpoint change)''. We use the official split with 67,660 for training, 3,976 for validation and 7,970 for testing.

\textbf{Spot-the-Diff} dataset \cite{DBLP:conf/emnlp/JhamtaniB18} contains 13,192 well-aligned image pairs from surveillance cameras. 
Based on the official split, the dataset is split into training, validation, and testing with a ratio of 8:1:1.

Following the state-of-the-art methods \cite{Park2019RobustCC,DBLP:conf/eccv/ShiYGJ020,DBLP:conf/acl/TuYLLGYY21}, we use five standard metrics to evaluate the quality of generated sentences, \emph{i.e.}, BLEU-4 \cite{papineni2002bleu}, METEOR \cite{banerjee2005meteor}, ROUGE-L \cite{lin2004rouge}, CIDEr \cite{vedantam2015cider} and SPICE \cite{DBLP:conf/eccv/AndersonFJG16}. We get all the results based on the Microsoft COCO evaluation server \cite{chen2015microsoft}. 

\subsection{Implementation Details}
We use ResNet-101 \cite{he2016deep} pre-trained on the Imagenet dataset \cite{russakovsky2015imagenet} to extract object features, with the dimension of 1024 $\times$ 14 $\times$ 14. We project these features into a lower dimension of 256. The hidden size of overall model is set to 512 and the number of attention heads in relation-embedded module is set to 4. The number of skeletons in an image pair is set to 50. The dimension of words is set to 300. For the hyper-parameter $\lambda$, we empirically set it as 0.1. In the training phase, we use Adam optimizer \cite{kingma2014adam} with the learning rate of 1 $\times$ $10^{-3}$, and set the mini-batch size as 128 and 64 on CLEVR-Change and Spot-the-Diff. At inference, for fair comparison, we follow the pioneer works \cite{Park2019RobustCC,DBLP:conf/emnlp/JhamtaniB18} in the two datasets to use greedy decoding strategy for caption generation. Both training and inference are implemented with PyTorch \cite{paszke2019pytorch} on a Tesla P100 GPU.

\begin{table*}[t]
\centering
\begin{threeparttable}
  \caption{Comparing with state-of-the-art methods on CLEVR-Change in Total Performance. RL refers to the training strategy of reinforcement learning.}
  \label{table3}
\begin{tabular}{c|c|c|c|c|c|c}
\hline
          & &\multicolumn{5}{c}{Total}                                                                \\ \hline
Method    & RL & B-4           & M             & R             & C              & S              \\ \hline
Capt-Dual \cite{Park2019RobustCC} &$\times$ & 43.5          & 32.7          & -             & 108.5           & 23.4                 \\ \hline
DUDA \cite{Park2019RobustCC}   &$\times$  & 47.3          & 33.9          & -             & 112.3          & 24.5                \\ \hline
M-VAM  \cite{DBLP:conf/eccv/ShiYGJ020}  &$\times$ & 50.3          & 37.0          & 69.7          & 114.9          & 30.5               \\ \hline
M-VAM+RAF \cite{DBLP:conf/eccv/ShiYGJ020} & \checkmark & 51.3          & 37.8          & 70.4          & 115.8          & 30.7           \\ \hline
R$^3$Net+SSP     & $\times$   & \textbf{54.7} & \textbf{39.8} & \textbf{73.1} & \textbf{123.0} & \textbf{32.6}       \\ \hline
\end{tabular}
\end{threeparttable}
\end{table*}

\begin{table*}[t]
\centering
\begin{threeparttable}
  \caption{Comparing with state-of-the-art methods on CLEVR-Change in terms of two settings.}
  \label{table4}
\begin{tabular}{c|c|c|c|c|c|c|c|c|c}
\hline
          & & \multicolumn{4}{c|}{Scene Change}           &                  \multicolumn{4}{c}{None-scene Change}                              \\ \hline
Method    & RL & B-4           & M             & C              & S             & B-4           & M             & C              & S              \\ \hline
Capt-Dual \cite{Park2019RobustCC} &$\times$ & 38.5          & 28.5          & 89.8           & 18.2           & 56.3          & 44.0          & 108.9           & 28.7              \\ \hline
DUDA  \cite{Park2019RobustCC}  &$\times$  & 42.9          & 29.7          & 94.6           & 19.9          & 59.8          & 45.2          & 110.8          & 29.1             \\ \hline
M-VAM+RAF \cite{DBLP:conf/eccv/ShiYGJ020} & \checkmark & -             & -             & -              & -             & -             & \textbf{66.4} & \textbf{122.6} & 33.4              \\ \hline
R$^3$Net+SSP    &$\times$    & \textbf{52.7} & \textbf{36.2} & \textbf{116.6} & \textbf{30.3} & \textbf{61.9} & 50.5          & 116.4          & \textbf{34.8}       \\ \hline
\end{tabular}
\end{threeparttable}
\end{table*}

\subsection{Ablation Studies}
To figure out the contribution of each module of the proposed network, we conduct the following ablation studies on CLEVR-Change: (1) Baseline which is based on DUDA \cite{Park2019RobustCC}; (2) RRM which is the  representation reconstruction module; (3) R$^3$Net which augments the RRM with a relation-embedded module; (4) R$^3$Net+SSP which augments the R$^3$Net with a syntactic skeleton predictor. 

\textbf{The evaluation on Total Performance.} Total performance is to simultaneously evaluate the model under both scene change and none-scene change. Experimental results are shown in Table \ref{table1}. We can observe that each module and the full method improve the total performance of Baseline. This indicates that our method not only can correctly judge whether there is an semantic change between a pair of images, but also can describe the change in an accurate natural language sentence.

\textbf{The evaluation on the settings of Scene Change and None-scene Change.} In the setting of scene change, both object and viewpoint changes happen. In the setting of none-scene change, there are only distractors, such as viewpoint change and illumination change. The experimental results are shown in Table \ref{table2}. Under the setting of scene change, we can observe that 1) the RRM, R$^3$Net, and R$^3$Net+SSP all significantly improve the Baseline; 2) the R$^3$Net is much better than the RRM; 3) the best performance is achieved when augmenting the R$^3$Net with the SSP. The above observations indicate that 1) compared to direct subtraction between a pair of unaligned images, it is effective to capture difference representation via the R$^3$Net, because it can overcome the distraction of viewpoint change; 2) learning semantic relations among object features is important, because these relations can enrich the raw object features, helpful for exploring fine-grained changes; 3) the SSP can enhance the semantic interaction between change localization and caption generation, and thus further improve the quality of generated sentences. 

Besides, under the setting of non-scene change, we can observe that the RRM is worse than the Baseline on some metrics. Our conjecture is that, on one hand, due to the large amount of clutter and only representing the image pair at feature-level, the RRM cannot learn the exact semantic similarities of corresponding locations in the two images, performing worse on some metrics. On the other hand, the Baseline learns a coarse difference representation between two unaligned images by a direct subtraction, so it is prone to learn a wrong change type or simply judge nothing has changed. This leads to the results that it performs worse than the RRM with the total performance and scene change, but achieves higher scores than the RRM on some metrics with none-scene change. In fact, when embedding semantic relations among object features, the R$^3$Net outperforms the Baseline in the both settings. This further indicates that it is beneficial to thoroughly understand image content via modeling semantic relations among object features.

\begin{table*}[t]
\centering
\begin{threeparttable}
  \caption{A Detailed breakdown of Change Captioning evaluation on CLEVR-Change by different change types: ``Color'' (C), ``Texture'' (T), ``Add'' (A), ``Drop'' (D), and ``Move'' (M).}
  \label{table5}
\begin{tabular}{c|c|c|c|c|c|c|c}
\hline
Method   & RL  & Metrics & C              & T              & A             & D              & M              \\ \hline
Capt-Dual \cite{Park2019RobustCC} &$\times$ & CIDEr   & 115.8          & 82.7           & 85.7          & 103.0          & 52.6                \\ 
DUDA \cite{Park2019RobustCC}   &$\times$  & CIDEr   & 120.4          & 86.7           & 108.3         & 103.4          & 56.4              \\ 
M-VAM+RAF \cite{DBLP:conf/eccv/ShiYGJ020} & \checkmark & CIDEr   & 122.1          & 98.7           & \textbf{126.3}         & 115.8          & 82.0              \\ 
R$^3$Net+SSP   &$\times$      & CIDEr   & \textbf{139.2} & \textbf{123.5} & 122.7         & \textbf{121.9} & \textbf{88.1}    \\ \hline
Capt-Dual \cite{Park2019RobustCC} &$\times$ & METEOR  & 32.1           & 26.7           & 29.5          & 31.7           & 22.4              \\ 
DUDA \cite{Park2019RobustCC}  &$\times$   & METEOR  & 32.8           & 27.3           & 33.4          & 31.4           & 23.5              \\ 
M-VAM+RAF \cite{DBLP:conf/eccv/ShiYGJ020} & \checkmark & METEOR  & 35.8           & 32.3           & 37.8          & 36.2           & 27.9              \\ 
R$^3$Net+SSP     &$\times$   & METEOR  & \textbf{38.9}  & \textbf{35.5}  & \textbf{38.0} & \textbf{37.5}  & \textbf{30.9}        \\ \hline
Capt-Dual \cite{Park2019RobustCC} &$\times$ & SPICE   & 19.8           & 17.6           & 16.9          & 21.9           & 14.7            \\ 
DUDA \cite{Park2019RobustCC} &$\times$    & SPICE   & 21.2           & 18.3           & 22.4          & 22.2           & 15.4              \\ 
M-VAM+RAF \cite{DBLP:conf/eccv/ShiYGJ020} & \checkmark & SPICE   & 28.0           & 26.7           & 30.8          & \textbf{32.3}           & 22.5          \\
R$^3$Net+SSP   &$\times$      & SPICE   & \textbf{31.6}  & \textbf{30.8}  & \textbf{32.3} & 31.7           & \textbf{25.4}        \\ \hline
\end{tabular}
\end{threeparttable}
\end{table*}

\begin{table}[h]
\centering
\begin{threeparttable}

  \caption{Comparing with state-of-the-art methods on Spot-the-Diff.}
  \label{table6}
\begin{tabular}{c|c|c|c|c|c}
\hline
Method      & RL    & M             & R             & C             & S             \\ \hline
DDLA    &$\times$        & 12.0          & 28.6          & 32.8          & -             \\ 
DUDA    &$\times$        & 11.8          & 29.1         & 32.5          & -             \\ 
SDCM    &$\times$        & 12.7          & 29.7          & 36.3          & -             \\
FCC    &$\times$        & 12.9          & 29.9          & 36.8          & -             \\
static rel-att &$\times$ & 13.0          & 28.3          & 34.0          & -             \\ 
dynamic rel-att &$\times$ & 12.2          & 31.4          & 35.3          & -             \\ 
M-VAM    &$\times$       & 12.4          & 31.3          & 38.1          & 14.0          \\ 
M-VAM+RAF  & \checkmark     & 12.9          & \textbf{33.2} & \textbf{42.5} & 17.1          \\ \hline
R$^3$Net+SSP    &$\times$           & \textbf{13.1} & 32.6          & 36.6          & \textbf{18.8} \\ \hline
\end{tabular}
\end{threeparttable}
\end{table}

\subsection{Performance Comparison}
\subsubsection{Results on CLEVR-Change Dataset}
In this dataset, we compare with four state-of-the-art methods, Capt-Dual \cite{Park2019RobustCC}, DUDA \cite{Park2019RobustCC}, M-VAM \cite{DBLP:conf/eccv/ShiYGJ020}, and M-VAM+RAF \cite{DBLP:conf/eccv/ShiYGJ020}, in four settings: 1) scene change; 2) none-scene change; 3) total (scene and none-scene changes); 4) specific types of scene change. 

From Table \ref{table3} and Table \ref{table4}, under two kinds of settings and total performance, we can observe that our method surpasses Capt-Dual and DUDA with a large margin. Compared to M-VAM+RAF, the total performance of our method is much better than it, which indicates our method is more robust. As shown in Table \ref{table4}, under the setting of none-scene change, it outperforms our method on METEOR and CIDEr. This could be a benefit of the reinforcement learning, while also sharply increasing the training time and computation complexity.

Table \ref{table5} is the specific change types. Among five changes, the most challenging types are `` Texture'' and ``Move'', because they are always confused with irrelevant illumination or viewpoint changes. Compared to the SOTA methods, our method achieves excellent performances under both change types. This shows that our method can better distinguish the attribute change or movement of objects from the illumination or viewpoint change. 

Hence, compared to the current SOTA methods from different dimensions, the generalization ability of our method is much better. This benefits from the merits that 1) the R$^3$Net can learn the fine-grained change and overcome viewpoint changes in the process of representation reconstruction; 2) the SSP can enhance the semantic interactions between change localization and caption generation.

\begin{figure}[t]
\centering
\includegraphics[width=0.8\textwidth]{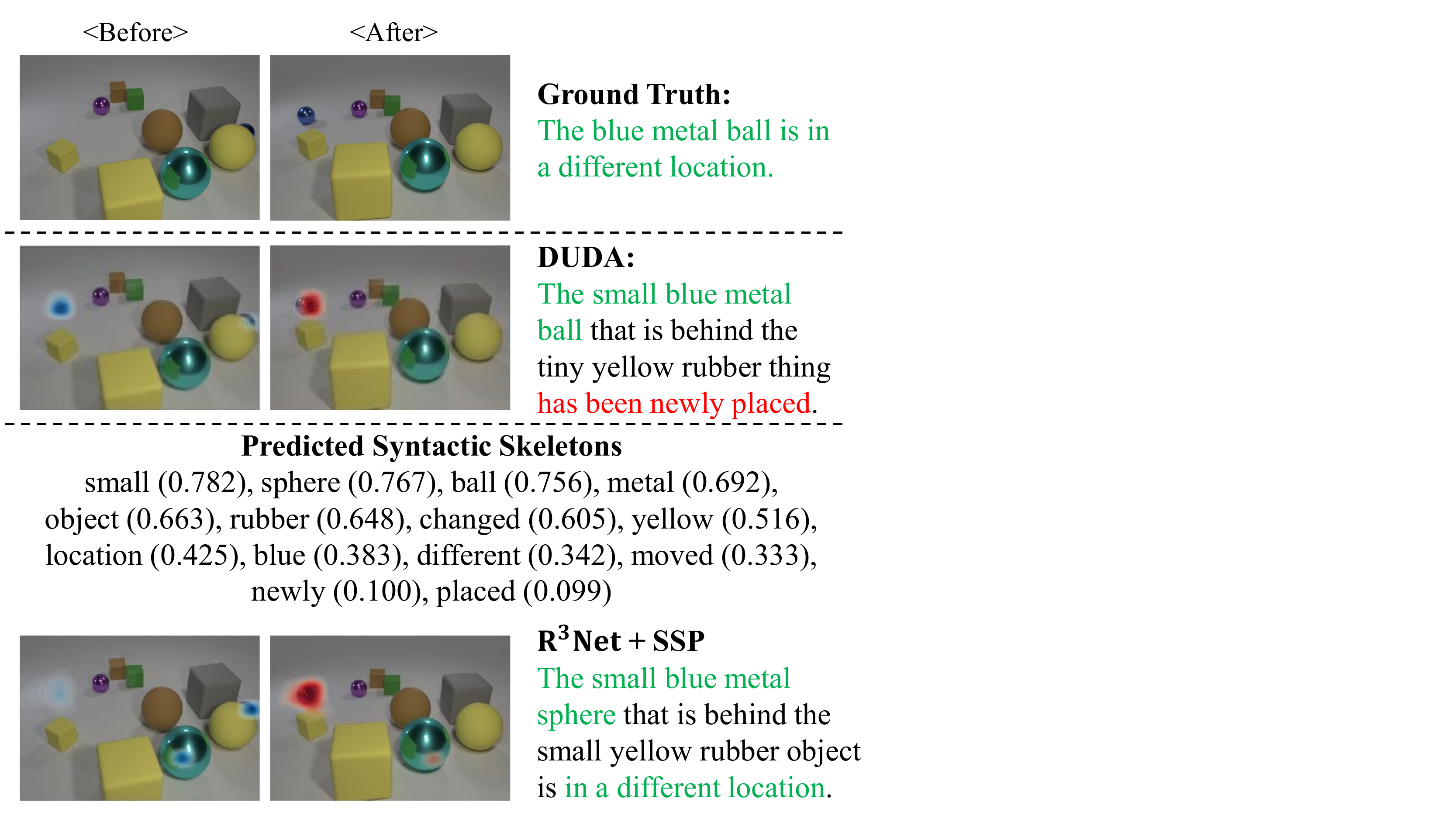} 
\caption{A example about ``Move'' case from the test set of CLEVR-Change, which involves the caption generated by humans (Ground Truth), DUDA (current SOTA method) and R$^3$Net+SSP. We also visualize the predicted syntactic skeletons and the localization results on the ``before'' (blue) and ``after'' (red). }
\label{fig3}
\end{figure}

\begin{figure*}[t]
\centering
\setlength{\abovecaptionskip}{-2.3cm}
\includegraphics[width=1\textwidth]{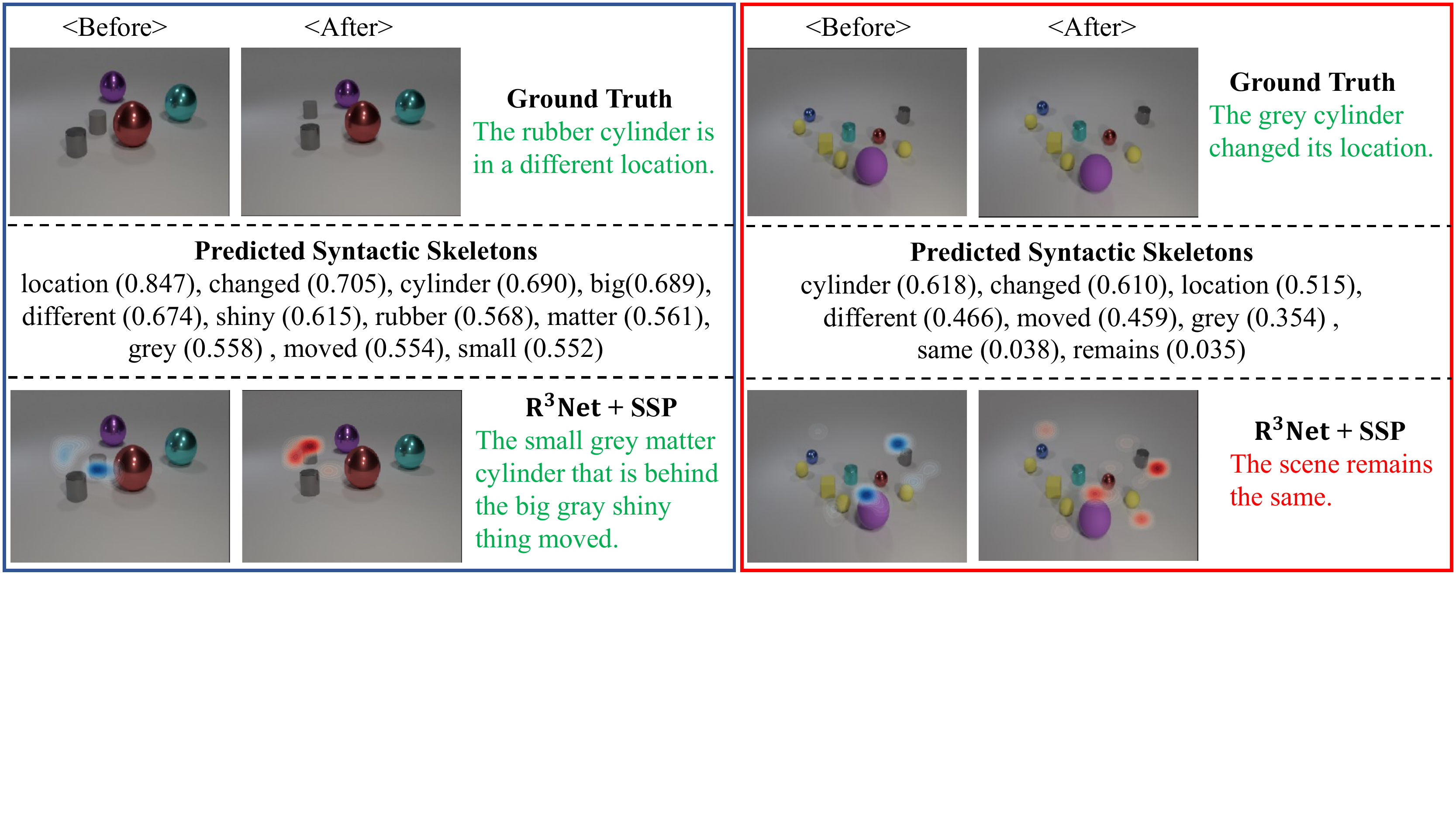} 
\caption{Qualitative examples of R$^3$Net+SSP. The left is a successful case that R$^3$Net+SSP localizes the accurate changed object and generates a correct sentence to describe the change. The right is a failure case that a slight movement of the object is not correctly described. }
\label{fig4}
\end{figure*}

\subsubsection{Results on Spot-the-Diff Dataset}
The image pairs in this dataset are mostly well aligned. We compare with eight SOTA methods and most of them cannot consider handling viewpoint changes: DDLA \cite{DBLP:conf/emnlp/JhamtaniB18}, DUDA \cite{Park2019RobustCC}, SDCM \cite{oluwasanmi2019captionnet}, FCC \cite{oluwasanmi2019fully}, static rel-att / dyanmic rel-att \cite{tan2019expressing}, and M-VAM / M-VAM+RAF \cite{DBLP:conf/eccv/ShiYGJ020}.

The results are shown in Table \ref{table6}. We can observe that when training without reinforcement learning, our method achieves the best performances on METEOR, ROUGE-L and SPICE. Compared to M-VAM+RAF trained by the reinforcement learning strategy, our method still outperforms them on METEOR and SPICE. Since there is no viewpoint change in this dataset, the superiority mainly results from that the relation-embedded module can enhance the fine-grained representation ability of object features, and the syntactic skeleton predictor can enhance the semantic interaction between change localization and caption generation.

\subsection{Qualitative Analysis}
Figure \ref{fig3} shows an example about the case of ``Move'' from the test set of CLEVR-Change. 
We can observe that DUDA localizes a wrong region on the ``before'' and thus misidentifies ``Move'' as ``Add''. By contrast, the R$^3$Net+SSP can accurately locate the moved object on the ``before'' and ``after'' images, which benefits from two merits. First, the R$^3$Net is able to localize the fine-grained change in the presence of viewpoint changes. Second, the SSP can predict the key skeletons based on the representations of image pair and their difference learned from the R$^3$Net. For instance, the skeletons of  ``changed'' and ``location'' has the higher probability scores than ``newly'' and ``placed''. This can provide the decoder with high-level semantic cues to generate the correct sentence.

Figure \ref{fig4} illustrates two cases about ``Move''. In the left example, the R$^3$Net+SSP successfully distinguishes the changed object (i.e., small grey cylinder) and predicts accurate skeletons with high probability scores. The right example is a failure case. In general,  we can observe that the grey cylinder is localized and the main skeletons are predicted, which indicates that the R$^3$Net learns a reliable representation of difference. However, the decoder still generates the wrong sentence. The reason behind the failure may be that the movement of this cylinder is very slight and the decoder receives the weak information of change (including skeletons). In our opinion, a possible solution for this challenge is to model position information for object features, which would enhance their position representation ability and help localize the slight movements.

\section{Conclusion}
In this paper, we propose a relation-embedded representation reconstruction network (R$^3$Net) and a syntactic skeleton predictor (SSP) to address change captioning in the presence of viewpoint changes, where the R$^3$Net can explicitly distinguish semantic changes from viewpoint changes and the SSP is to enhance the semantic interaction between change localization and caption generation. Extensive experiments show that the state-of-the-art results are achieved on two public datasets, CLEVR-Change and Spot-the-Diff.



\section*{Acknowledgements}
The work was supported by National Natural Science Foundation of China (No. 61761026, 61972186, 61732005, 61762056, 61771457, and 61732007), 
in part by Youth Innovation Promotion Association of Chinese Academy of Sciences (No. 2020108), and CCF-Baidu Open Fund (No. 2021PP15002000).

\bibliography{anthology,custom}
\bibliographystyle{acl_natbib}
\end{document}